\documentclass[conference]{IEEEtran}

\IEEEoverridecommandlockouts                              

\usepackage{times}
\usepackage{helvet}  
\usepackage{courier}  
\usepackage{url}  
\usepackage{graphicx}  
\usepackage{amsmath} 

\usepackage{amssymb} 
\usepackage{nicefrac}
\usepackage{bm}
\usepackage{siunitx}
\usepackage{soul}
\usepackage{graphics} 
\usepackage{epsfig} 
\usepackage{epstopdf}
\graphicspath{ {figs} {imgs}}
\usepackage{caption}
\usepackage{xcolor}
\usepackage{subcaption}
\usepackage{lipsum}
\usepackage{float}
\usepackage{todonotes}
\usepackage{color}
\usepackage{soul}
\usepackage{stfloats}

\usepackage[numbers]{natbib}
\usepackage{multicol}
\usepackage[bookmarks=true]{hyperref}

\newcommand{\bu}{\mathbf{u}}
\newcommand{\bv}{\mathbf{v}}

\newcommand{\ts}{\textsuperscript}

\begin{document}

\title{Neural SLAM: \\Learning to Explore with External Memory}

\author{
  Jingwei Zhang$^{1}$ \
  Lei Tai$^{2}$ \
  Ming Liu$^{2}$ \
  Joschka Boedecker$^{1}$ \
  Wolfram Burgard$^{1}$ \
  \thanks{$^{1}$Department of Computer Science, University of Freiburg. {\tt\small\{zhang, jboedeck, burgard\}@cs.uni-freiburg.de}}
  \thanks{$^{2}$Department of Electronic and Computer Engineering, The Hong Kong University of Science and Technology. {\tt\small\{ltai, eelium\}@ust.hk}}
}

\maketitle

\begin{abstract}
We present a learning-based approach for agents to explore and cover unknown environments based on sensor data.
We achieve this by interacting reinforcement learning agents with external memory architectures, in which the external memory acts as an internal representation of the environment for the agent. We embed procedures mimicking that of traditional simultaneous localization and mapping inside a completely differentiable deep neural work, which encourages the evolution of cognitive mapping behaviors during the process of learning.
We show that this approach can help reinforcement learning agents to successfully explore and cover new environments, where long-term memory is essential for making informed planning decisions.
We validate our approach in both challenging grid-world environments and preliminary Gazebo experiments.
A video of our experiments can be found at:
\url{https://goo.gl/G2Vu5y}.

\end{abstract}
\IEEEpeerreviewmaketitle

\section{Introduction}
\label{sec:introduction}

\subsection{Cognitive Mapping}

Studies of animal navigation have shown that the hippocampus plays an important role \cite{o1978hippocampus} \cite{mcnaughton2006path} \cite{collett2004animal}. It performs cognitive mapping that combines path integration and visual landmarks, so as to give the animals sophisticated navigation capabilities instead of just reflexive behaviors based only upon the immediate information they perceive.

Similarly, to successfully navigate and explore new environments in a timely fashion, intelligent agents would benefit from having their own internal representation of the environment whilst traverse, so as to go beyond the scope of performing reactive actions based on the most recent sensory input. Traditional methods in robotics thus developed a series of methods like simultaneous localization and mapping (SLAM), localization in a given map, path planning and motion control, to enable robots to complete such challenging tasks \cite{Thrun05} \cite{Lav06} \cite{Lat91}. Those individual components have been well studied and understood as separate parts, but here we view them as a unified system and attempt to embed SLAM-like procedures into a neural network such that SLAM-like behaviors maybe be able to evolve out of the course of reinforcement learning agents exploring new environments. This guided learned system could then benefit from each individual component (localization, mapping and planning) adapting in the awareness of each other's existence, instead of being rigidly combined together as in traditional methods. Also, in this paper we represent this system using a completely differentiable deep neural network, ensuring the learned representation is distributed and feature-rich, a property that rarely comes with traditional methods but is key to robust and adaptive systems \cite{bengio2013deep}. 
Several recent works \cite{kanitscheider2017training,gupta2017unifying} also seeks to cultivate the cognitive mapping behavior in learning agents.

\subsection{External Memory}

The memory structure in traditional recurrent neural networks (RNNs) like long short term memory networks (LSTMs) are ultimately short-term,
which would not be sufficient for developing informative navigation or exploration strategies. For the network to have an internal representation of the environment, i.e., its own cognitive map, an external memory architecture \cite{graves2014neural} \cite{graves2016hybrid} is required. Having an external memory, such as the neural turning machine (NTM) \cite{graves2014neural} or the differentiable neural computer (DNC) \cite{graves2016hybrid} besides a deep network separates the learning of computation algorithms from the storage of long-term memory. This is essential for learning successful exploration strategies, since if the computation and the memory are mixed together in the weights of the network, then with the memory demands increasing over time, the expressive capacity of the network would be very likely to decrease \cite{graves2016hybrid}.

Besides NTM and DNC, there is another branch of work on external memory architectures for deep networks which studies the memory networks. But the memory networks as in \cite{oh2016control} \cite{sukhbaatar2015end} do not learn what to write to the memory, which is not sufficient for our task because the network is expected to learn to map onto its external memory to aid planning.

The Neural Map as proposed in \cite{parisotto2017neural} adapted the $1D$ external memory in \cite{graves2014neural} to $2D$ as a form of structured map storage for an agent to learn to navigate. However, they do not utilize the $2D$  structure of this memory as all their operations can be conducted as if the memory address were a $1D$ vector. Furthermore, they assume the location of the agent is always known so as to write exactly to the corresponding location in the memory while the agent travels through the maze, a prerequisite that can rarely be met in real-world scenarios.

In our work, we also adopt the NTM approach for accessing external memory, since its behavior of writing to continuous blocks with its location focus is a natural fit for integrating the motion model, and corresponds well to SLAM operations. It is not straightforward to see as good an integration of DNC, as although its temporal link matrix can track down the writing operations, the dynamic nature of its memory allocation makes it not directly clear as how to embed motion.

\subsection{Embedding Classic Models into Deep Neural Networks}

Embedding domain-specific structures as inductive bias into neural networks can be found in many works.
Unlike methods that treat the networks completely as black-box approximators thus cannot benefit from the valuable prior knowledge accumulated over the years (it is like forcing a boy to deduct all the laws of physics from his observations by himself but not giving him the physics textbook to learn from), this line of formulation biases deep models toward learning representations containing the structures that we already know they would benefit from having for specific domains.

\cite{tamar2016value} embedded the value iteration procedures into a single network, forcing the network to learn representations following the well-defined policy-evaluation, policy-improvement loop, while benefiting from the feature-rich representations from deep architectures.
\cite{gupta2017cognitive} went one step further by using the value iteration network as the planning module inside a visual navigation system. They treat an internal part of the network as an egocentric map and apply motion on it.
\cite{fischer2015flownet} added a cross-correlation layer to compute correlations of features of corresponding neighboring cells between subsequent frames, which explicitly provides the network with matching capabilities. This greatly helps the learning of optical flow since the optical flow is computed based on local pixel dynamics.
\cite{zhang2016deep} forced the network to learn representative features across tasks by explicitly embedding structures mimicking the computation procedures of successor feature reinforcement learning into the network, and their resulting architecture is able to transfer navigation policies across similar environments.
Also, more related to ours, \protect\cite{chaplot2018active} incorportates the filtering process of traditional localization into deep network architectures, and is able to train agents that can localize accurately and efficiently.

Traditionally, when using well-established models in a combined system with other modules, they typically do not benefit from the other components. This is because their behaviors can not adapt accordingly, as those models come out of deduction but have not evolved out of learning (directly applying those well established traditional models is like to directly give the boy all the answers to his physics questions instead of giving him the physics textbook for him to learn to solve those questions).  While if those functionalities are learned along with other components, their behaviors can influence each other and the system could potentially obtain performance beyond directly combining well-established models.

Let us take SLAM as an example. SLAM is used as a building block for complicated autonomous systems to aid navigation and exploration, yet the SLAM model and the path planning algorithms are individually developed, not taking each other into account. \cite{bhatti2016playing} augmented the state space of their reinforcement learning agent with the output of a traditional SLAM algorithm. Although this improves the navigation performance of the agent, it still experiences the issues discussed above since SLAM is rigidly placed into their architecture. While if SLAM-like behaviors can be encouraged to evolve out of the process of agents learning to navigate or to explore, then the resulting system would be much more deeply integrated as a whole, with each individual component influencing each other while benefiting from learning alongside each other. The SLAM model from the resulting system would evolve out of the need for exploration or navigation, not purely just for performing SLAM. Additionally, if learning with deep neural nets, the resulting models will be naturally feature-rich, which is rarely a property of traditional well-established models.

Although a number of works have been presented on utilizing deep reinforcement learning algorithms for autonomous navigation \cite{mirowski2016learning} \cite{zhu2016target} \cite{zhang2016deep} \cite{gupta2017cognitive} \cite{tai2017virtual}, none of them have an explicit external memory architecture to equip the agent with the capability of making long-term decisions based on an internal representation of a global map. Also, these works mainly focus on learning to navigate to a target location, while here we attempt to solve a more challenging task of learning to explore new environments under a time constraint, in which an effective long term memory mechanism is essential.

Following these observations, we attempt to embed the motion prediction step and the measurement update step of SLAM into our network architecture, by utilizing the soft attention based addressing mechanism in \cite{graves2014neural}, biasing the write/read operations towards traditional SLAM procedures and treating the external memory as an internal representation of the map of the environment, and train this model using deep reinforcement learning algorithms, to encourage the evolution of SLAM-like behaviors during the course of exploration.

However, we want to emphasize that, rather than learning to do occupancy grid mapping to provide metric localization solutions, we aim to find an architecture well suited to encourage the agent to evolve its own cognitive mapping.
We do not constrain the learned internal map to correspond one-to-one to an occupancy grid map;
instead,
since the learned map is to be used as an internal representation of the global environment,
it is down to the agent to learn to write what it deems as useful,
and what information is most relevant to read to make planning decisions.
The “localization” here extracts the center of mass of the access weights but not the real metric localization of the agent in the real world,
which is inaccessible in real scenarios.
Since all the procedures are learned end-to-end,
those operations
(rigidly combined together in traditional methods)
are learned together to benefit each other.

\subsection{Exploration and Coverage in Unknown Environments}
Effective exploration capabilities are required of intelligent agents to perform tasks like floor sweeping, surveillance, rescue and sample collection \cite{shen2012autonomous}.
Especially for
Traditional techniques for exploration includes information gain based approaches, goal assignment using coverage maps or occupancy grid maps, etc \cite{stachniss2009robotic}. However, such techniques require building and maintaining accurate maps of the environment for the agent to memorize the already explored areas, in which loop closure plays an important role.

\cite{mirowski2016learning} added loop closure detection, along with depth prediction as auxiliary tasks to provide additional supervision signals when training reinforcement learning agents in environments with only sparse rewards. Specifically, the loop closure detection is trained via supervised learning by integrating the ground truth velocities of the agent, which is not accessible in real world scenarios. In this paper, we highlight the difference to \cite{mirowski2016learning} that the loop closure is learned implicitly within our model via an embedded SLAM structure. Our strategy requires less input information and depends less on ground-truth information as supervision. Additionally, we tested our approach in the Gazebo environment \cite{koenig2004design} which is more realistic with respect to the underlying physics and the sensor noise compared to the simulated environment used in \cite{mirowski2016learning}. Additionally, compared with traditional SLAM-based methods, our strategy eliminates the need for building and maintaining an expensive map for each new environment and only needs a forward pass through the trained model to give out planning decisions, which runs $200\si{\hertz}$ on CPU. This enables our agent to cope with the limited memory and processing capabilities on robotics platforms.

\section{Methods}
\label{sec:methods}

\subsection{Background}
\label{sec:mdp}

\begin{figure*}[!t]
   \centering
   \includegraphics[width=\textwidth]{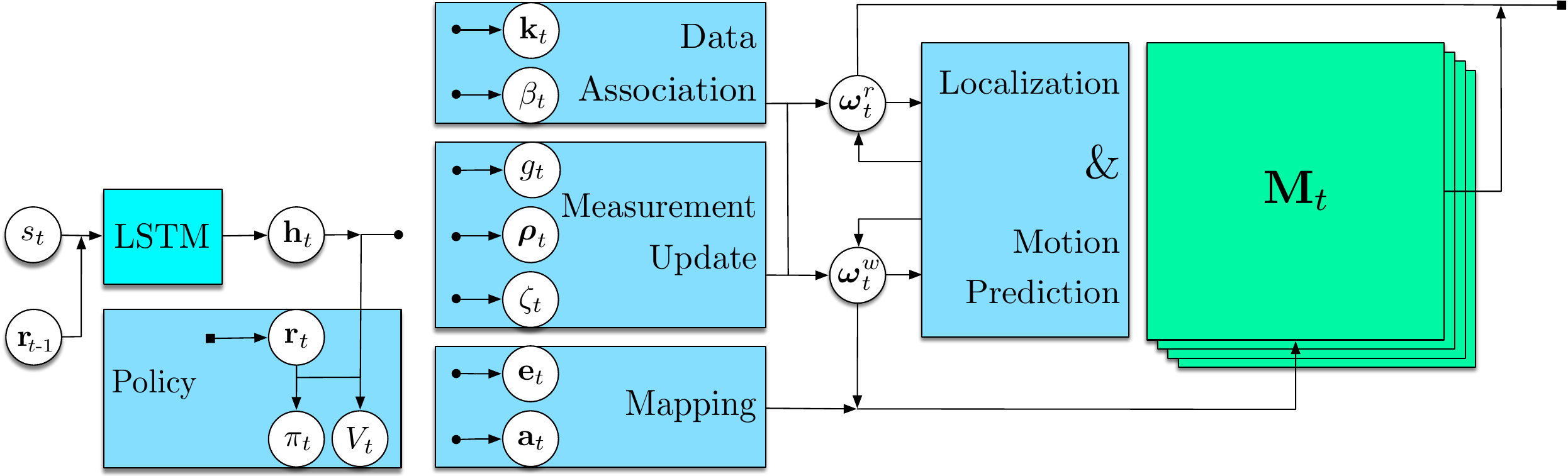}
   \caption{Visualization of the Neural-SLAM model architecture (we intentionally use \textit{blue} for the components in charge of \textit{computation}, \textit{green} for \textit{memory}, and \textit{cyan} for a mixture of both.)}
   \label{fig:model}
\end{figure*}

We formulate the exploration task as a Markov decision process (MDP) in which the agent interacts with the environment through a sequence of observations, actions and rewards.
At time step $t \in [0, T]$,
the agent resides at state $S_t=s_t\in\mathcal{S}$.
It then selects an action $A_{t}=a_t\in\mathcal{A}$ based on the policy
$\pi(\cdot|s_t)$,
which corresponds to a motion command
for the agent to execute.
By taking the action,
the agent receives a scalar reward signal $R_{t+1}\in\mathbb{R}$ and transits to the next state $S_{t+1}$.
The goal for the agent is to maximize the expected return,
and the return is defined as the discounted cumulative future reward ($\gamma \in (0,1]$ is the discount factor):
$
G_t
=
\sum_{\tau=t+1}^{T}\gamma^{\tau-t-1}R_{\tau}
$.

In this paper we build upon the asynchronous advantage actor-critic algorithm (A3C) \cite{mnih2016asynchronous} as the backbone deep reinforcement learning method,
together with the truncated generalized advantage estimator (GAE) \cite{schulman2015high}
to learn optimal policies.
More specifically,
in A3C multiple parallel workers are deployed to collect rollouts (each of up to a horizon of $K$) by interacting with each of their own copies of the environment.
Since typically $K$ is smaller than $T$,
a multi-step notion of the return is defined as:
$
G_{t:K}
=
\sum_{\tau=t+1}^{K}\gamma^{\tau-t-1}R_{\tau}
+
\gamma^{K}V(S_K)
$,
where the future reward beyond the horizon of $K$ is bootstrapped by its estimate $V(S_K)$,
with
$
V(s_t)
=
\mathbb{E}_{\pi}
[
G_t
|
S_t=s_t
]
$.
A $k$-step advantage is defined as
\begin{align}
    A^{(t:t+k)}
&=
    G_{t:t+k}
    -
    V(S_t)
=
    \sum_{\tau=0}^{k-1}
    \gamma^{\tau} \delta_{t+\tau}^V,
\end{align}
where
$
\delta_{t}^V
=
R_{t+1}
+
\gamma V(S_{t+1})
-
V(S_{t})
$
is the Bellman residual.
The truncated generalized advantage estimator GAE \cite{schulman2015high}
with
a horizon of $K$ is then defined as
$
    \hat{A}^{(t:K)}
=
    \sum_{\tau=0}^{K-1-t}
    (\gamma\lambda)^{\tau} \delta_{t+\tau}^V
$,
an exponentially weighted average
(by $\lambda$) of those multi-step advantage estimates.
In A3C,
the value function network is parameterized by $\bm{\theta}_V$.
In the meanwhile,
it also maintains a policy network with parameters
$\bm{\theta}_{\pi}$.
Each worker calculates gradients with its collected experiences with the following equations
\cite{mnih2016asynchronous}
\begin{align}
    d\bm{\theta}_{\pi}
=&
    \nabla_{\bm{\theta}_{\pi}} \log \pi(a_{t}|s_{t}; \bm{\theta}_{\pi})
    \hat{A}^{(K)}(s_t,a_t;\bm{\theta}_V)
\notag\\
 &+ \alpha_{\text{ent}} \nabla_{\bm{\theta}_{\pi}}
 H(\pi(a_{t}|s_{t};\bm{\theta}_{\pi})),
\label{equ:policygrad}\\
    d\bm{\theta}_{V}
=&
    \partial(Y_{t} - V(s_{t};\bm{\theta}_{V}))^{2} / \partial\bm{\theta}_{V},
\label{equ:valuegrad}
\end{align}
where
$H$ denotes the entropy and $\alpha_{\text{ent}}$ is the weight of the entropy regularization,
$
Y_t
=
V(s_t;\bm{\theta}_V)
+
\hat{A}^{(K)}(s_t,a_t;\bm{\theta}_V)
$
(the dependency of $Y_t$ on $\bm{\theta}_V$ is intentionally left out in its notation as $Y_t$ is treated as a plain target value but not as a function of the learnable parameter in the gradient calculations).
These gradients are used to asynchronously update a shared model in the HOGWILD! fashion
\cite{recht2011hogwild}.

\subsection{Neural SLAM Architecture}

As discussed previously, we require our model to have an external memory structure for the agent to utilize as an internal representation of the environment. Thus, we added an external memory chunk
$\mathbf{M}$
of size $H \times W \times C$ (containing $H \times W$ memory slots, with $C$ channels or features for each slot), which can be accessed by the network via a write head and a read head. (We note that our work can be easily extended to multiple heads for write/read, but in this paper we only investigate with one write/read head. We also observe that the number of heads can be viewed as the number of particles as in particle filter \cite{Thrun05}.)

At each time step $t$,
the input is fed directly to an LSTM cell, which gives out a hidden state $\mathbf{h}_{t}$.
Following the design principles of the neural turing machine \cite{graves2014neural},
this hidden state $\mathbf{h}_{t}$ is then used in each head to emit a set of control variables
\{$\mathbf{k}_{t}$, $\beta_{t}$, $g_{t}$, $\bm{\rho}_{t}$, $\zeta_{t}$\}
(each write head additionally emits
\{$\mathbf{e}_{t}$, $\mathbf{a}_{t}$\}
)
through a set of linear layers. The write head and the read head then each computes their access weight
($\bm{\omega}_{t}^{w}$ and
$\bm{\omega}_{t}^{r}$,
both of size $H \times W$) based on those control variables.
Then the write head would use its access weight
$\bm{\omega}_{t}^{w}$ along with
$\bm{e}_{t}$ and
$\bm{a}_{t}$ to write to the memory
$\mathbf{M}_{t-1}$,
while the read head would access the updated memory
$\mathbf{M}_{t}$,
with its access weight
$\bm{\omega}_{t}^{r}$
to output a read vector
$\mathbf{r}_{t}$.
Next,
$\mathbf{h}_{t}$ and
$\mathbf{r}_{t}$ are concatenated together to compute the final output:
a policy distribution
$\pi_{t}$,
and a value estimate $V_{t}$,
which are then used to calculate gradients to update the whole model according to
Eq.\ref{equ:policygrad} and
Eq.\ref{equ:valuegrad}.

The Neural-SLAM Model Architecture is shown in Fig.\ref{fig:model}.
We will describe the operations in each component in detail in the following section.

\subsection{Embedded SLAM Structure}
\label{sec:embeddedslam}
We use the same addressing mechanism for computing the access weights of the write head
$\bm{\omega}_{t}^{w}$ and the read head
$\bm{\omega}_{t}^{r}$,
except that the read head addressing happens after the write head updates the external memory, thus it would access the memory of the current time step. Below we describe the computations in detail, where we refer to both access weights at time step $t$ as
$\bm{\omega}_{t}$.

\subsubsection{Prior Belief}
\label{sec:prior}

We view the access weights of the heads as their current beliefs. We make the assumption that the initial pose of the agent is known at the beginning of each episode. Also,
the sensing range of its onboard sensor is known a priori. Then we initialize the access weight
$\bm{\omega}_{0}$
with a Gaussian kernel centering around the initial pose, filling up the whole sensing area and summing up to 1; all other areas are assigned with weight $0$. 
The external memory is initialized as
$\mathbf{M}_{0}=0$.

\subsubsection{Localization \& Motion Prediction}
\label{sec:motion}

At each time step $t$,
we first do a motion prediction,
by applying the motion command the agent receives
from the last time step
onto its access weight from the last time step
($f_{\text{mot}}$ here can be any motion model)
\begin{align}
    \bar{\bm{\omega}}_{t}
&=
    f_{\text{mot}}(\bm{\omega}_{t-1},
    a_{t-1}).
\end{align}

Note that since we view the external memory not as an egocentric map but as a global map, we need to first localize on the access weight before the motion model can be applied. Thus, we localize by first identifying the center of mass in the current access weight matrix as the position of the agent, then choose the direction with the largest sum of weights within the corresponding sensing area as its orientation.

\subsubsection{Data Association}
\label{sec:data}

Each head emits a key vector
$\mathbf{k}_{t}$
of length $C$,
which is compared with each slot
$(x,y)$
in the external memory
$\mathbf{M}_{t}(x,y)$ under a similarity measure
$f_{\text{sim}}$
(in this paper we use cosine similarity as in Eq.\ref{equ:cosine}),
to compute a content-based access weight
$\bm{\omega}_{t}^{c}$ based on the data association score
(each head also outputs a key strength scalar $\beta_{t}$ to increase or decrease the focus of attention)
\begin{gather}
    \bm{\omega}_{t}^{c}(x,y)
=
    \frac{\exp(\beta_{t}\cdot
        {f_{\text{sim}}(
            \mathbf{k}_{t},
            \mathbf{M}_{t}{(x,y)}
        ))}}
    {\sum_{(i,j)}\exp{(\beta_{t}\cdot
        f_{\text{sim}}(
            \mathbf{k}_{t},
            \mathbf{M}_{t}{(i,j)}
        ))}},
\\
    f_{\text{sim}}(\bu, \bv)
=
    \frac{\bu \cdot \bv}{\lVert \bu \rVert \cdot \lVert \bv \rVert }.
\label{equ:cosine}
\end{gather}

\subsubsection{Measurement Update}
\label{sec:measurement}

We then perform a measurement update with the following steps.

First, the content-based access weight from this time step
$\bm{\omega}_{t}^{c}$
and the last access weight after motion prediction $\bar{\bm{\omega}}_{t}$
are interpolated together using an interpolation gate scalar
$g_{t}$ generated by each head
\begin{align}
    \bm{\omega}_{t}^{g}
=
    g_{t} \bm{\omega}_{t}^{c} + (1 - g_{t}) \bar{\bm{\omega}}_{t}.
\end{align}

Then, a shift operation is applied based on the shift kernel
$\bm{\rho}_{t}$
emitted by each head
(in this paper
$\bm{\rho}$
defines a normalized distribution over a $3\times3$ area), to account for the noise in motion and measurement.
This shift operation can be viewed as a convolution over the access weight matrices, with
$\bm{\rho}_{t}$
being the convolution kernel
\begin{align}
    \bm{\omega}_{t}^\rho(x,y)
&=
    \sum_{i=0}^{H-1}
    \sum_{j=0}^{W-1}
    \bm{\omega}_{t}^{g}(i,j)\bm{\rho}_{t}(x-i, y-j).
\end{align}

Finally, the smoothing effect of the shift operation is compensated with a sharpen scalar
$\zeta_{t} >= 1$
\begin{align}
      \bm{\omega}_{t}{(x,y)}
&=
      \frac{{\bm{\omega}_{t}^{\rho}(x,y)}^{\zeta_t}}
      {\sum_{(i,j)}{\bm{\omega}_{t}^{\rho}(i,j)}^{\zeta_t}}
\end{align}

\subsubsection{Mapping}
\label{sec:mapping}

The write head each generates two additional vectors (each contains $C$ elements):
an erase vector
$\mathbf{e}_{t}$ and an add vector
$\mathbf{a}_{t}$.
Along with its access weight
$\bm{\omega}_{t}^{w}$,
the write head accesses and updates the external memory with the following operations
\begin{align}
    \tilde{\mathbf{M}}_{t}
&=
    \mathbf{M}_{t-1}
    (1 - \bm{\omega}_{t}^{w} \mathbf{e}_{t}),
\\
    \mathbf{M}_{t}
&=
    \tilde{\mathbf{M}}_{t} + \bm{\omega}_{t}^{w} \mathbf{a}_{t}.
\end{align}
\subsection{Policy}
\label{sec:planning}

After the memory has been updated to
$\mathbf{M}_{t}$,
it is accessed by the read head by its access weight
$\bm{\omega}_{t}^{r}$,
to output a read vector
$\mathbf{r}_{t}$
(which can be seen as a summary of the current internal map representation)
\begin{align}
    \mathbf{r}_{t}
&=
    \sum_{(i,j)} \bm{\omega}_{t}^{r}(i,j)
    \mathbf{M}_{t}{(i,j)}.
\end{align}

This read vector
$\mathbf{r}_{t}$
is then concatenated with the hidden state
$\mathbf{h}_{t}$,
and fed into two linear layers
$\bm{\phi}_{\pi}$ and
$\bm{\phi}_{V}$
(we note that the policy network
$\bm{\theta}_{\pi}$ and
the value network
$\bm{\theta}_{V}$
share parameters except for their output layers,
which are parameterized by
$\bm{\phi}_{\pi}$ and
$\bm{\phi}_{V}$ respectively)
to give out the policy distribution and the value estimate
\begin{align}
    \pi_{t}
&=
    \text{softmax}(
        \bm{\phi}_{\pi}([
            \mathbf{h}_{t},
            \mathbf{r}_{t}
        ])),
\\
    V_{t}
&=
    \bm{\phi}_{V}([\mathbf{h}_{t}, \mathbf{r}_{t}]).
\end{align}

$\pi_{t}$ and $V_{t}$ are subsequently used for calculating losses for on-policy deep reinforcement learning,
as discussed in Sec.\ref{sec:mdp}.
An action $a_{t}$ is then drawn from a multinomial distribution defined by $\pi_{t}$ during training,
while a greedy action is taken during evaluation and testing.

We refer to Sec.\ref{sec:setup} for a detailed description of the reward function used in this paper,
as well as a brief discussion for the possibility of extracting intrinsic rewards out of the external memory.

\section{Experiments}
\label{sec:experiments}

\subsection{Experimental Setup}
\label{sec:setup}

\begin{figure}[b]
    \centering
        \includegraphics[height=0.8in]{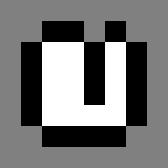}
\vspace{0.1cm}
        \includegraphics[height=0.8in]{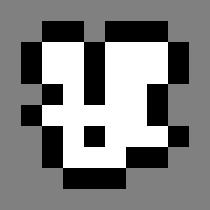}
\vspace{0.1cm}
        \includegraphics[height=0.8in]{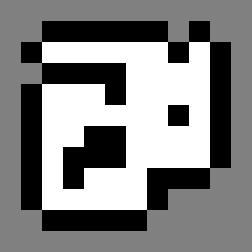}
\\
        \includegraphics[height=0.8in]{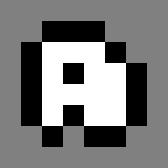}
\vspace{0.1cm}
        \includegraphics[height=0.8in]{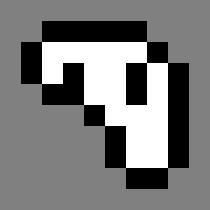}
\vspace{0.1cm}
        \includegraphics[height=0.8in]{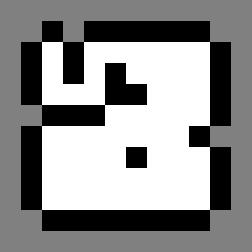}
    \caption{Visualization of randomly generated grid worlds,
    with sizes ranging from $8\times8$ (left) to $12\times12$ (right).}
    \label{fig:maps}
\end{figure}

\begin{figure}[!t]
    \centering
        \includegraphics[height=0.76in]{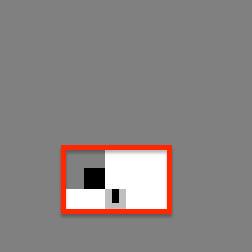}
        \vspace{0.1cm}
        \includegraphics[height=0.76in]{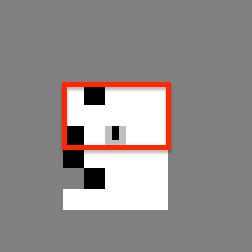}
        \vspace{0.1cm}
        \includegraphics[height=0.76in]{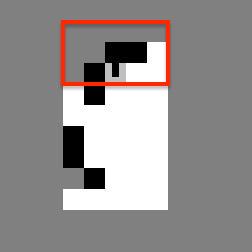}
        \vspace{0.1cm}
        \includegraphics[height=0.76in]{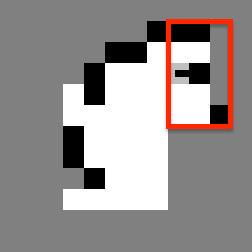}
        \\
        \includegraphics[height=0.76in]{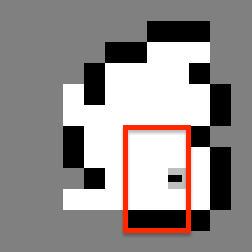}
        \vspace{0.1cm}
        \includegraphics[height=0.76in]{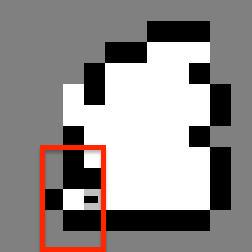}
        \vspace{0.1cm}
        \includegraphics[height=0.76in]{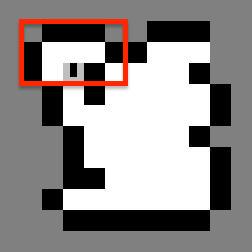}
        \vspace{0.1cm}
        \includegraphics[height=0.76in]{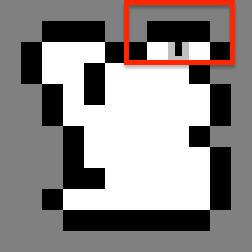}
    \caption{Visualization of a sample trajectory of a trained \textit{Neural-SLAM} agent successfully completing exploration and coverage in a new environment.
    The agent is visualized as a grey grid with a black rectangle pointing at its current orientation from its center.
    The obstacles are shown as black grids,
    free space as white grids,
    and grey grids indicate unexplored areas.
    The world clears up as the agent explores with its sensor
    (the sensor cannot see through walls nor across sharp angles),
    whose sensing area is shown as red bounding boxes
    (the information in the red bounding box is the observation received by the agent that is fed into the network:
    a $3\times5$ vector with its entries containing $1$ for black grids, $0$ for white and $0.5$ for grey).
    An exploration is completed when the agent has cleared up all possible grids,
    in which case the current episode is considered to be terminated and solved.
    An episode would also be terminated
    (but not considered as solved)
    when a maximum step of $750$ is reached.}
    \label{fig:demo}
\end{figure}

\begin{figure}[h]
    \centering
    \begin{subfigure}[t]{0.96\columnwidth}
        \centering
        \includegraphics[width=\textwidth]{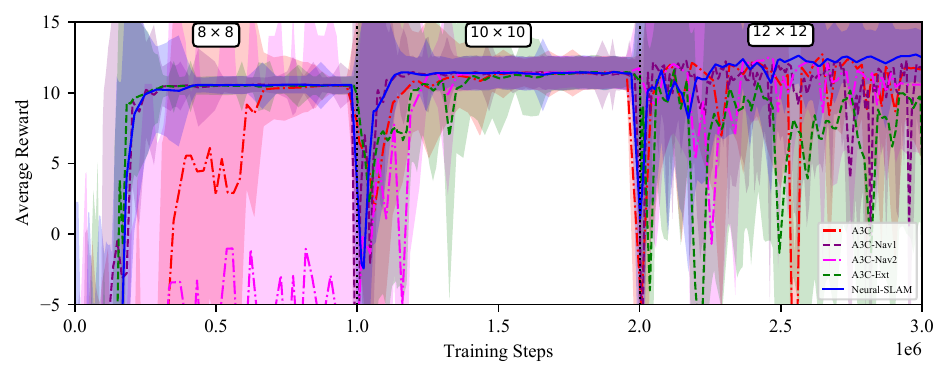}
    \end{subfigure}
    \caption{Comparison between the average reward obtained and number of episodes solved in $3000$ steps during evaluation by an \textit{A3C} agent (with 1 LSTM, motion command directly concatenated into the input), an \textit{A3C-Nav1} agent (with 2 stacked LSTMs, motion command directly concatenated into the input), an \textit{A3C-Nav2} agent (with 2 stacked LSTMs, motion command concatenated with the output of the $1\ts{st}$ LSTM, then input into the $2\ts{st}$ LSTM) , an \textit{A3C-Ext} agent (with 1 LSTM and an external memory, motion command concatenated with the output of the LSTM then fed to the external memory architecture, which is like the \textit{Neural-SLAM} without the Localization $\&$ Motion Prediction step, and our \textit{Neural-SLAM} agent (incorportate motion command with an explicit motion model, as discussed in Sec. \ref{sec:motion}).
    Each plot shows the mean over $3$ random seeds,
    the standard deviation is omitted for the right plot for a clearer visualization.
    We train continuously for $3$ courses transitioning from world sizes of $8\times8$ to $12\times12$.}
    \label{fig:curriculum}
\end{figure}

\begin{figure}[t]
    \centering
    \begin{subfigure}{0.053\textwidth}
    \centering
        \frame{\includegraphics[height=0.4in]{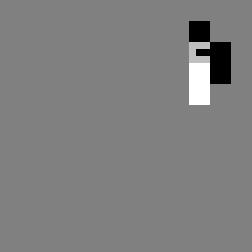}} \\
        \frame{\includegraphics[height=0.4in]{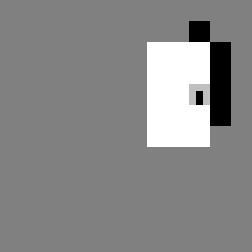}} \\
        \frame{\includegraphics[height=0.4in]{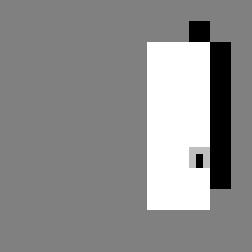}} \\
        \frame{\includegraphics[height=0.4in]{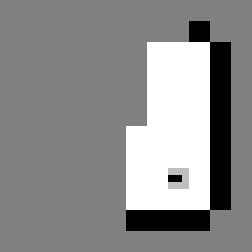}} \\
        \frame{\includegraphics[height=0.4in]{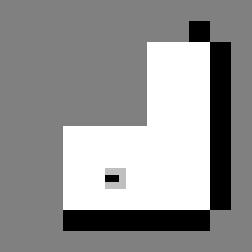}} \\
        \frame{\includegraphics[height=0.4in]{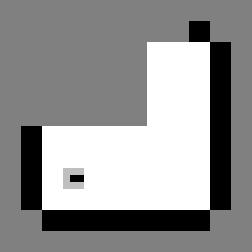}} \\
        \frame{\includegraphics[height=0.4in]{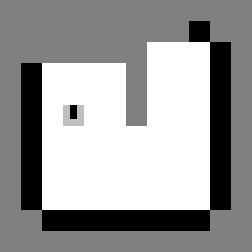}} \\
        \frame{\includegraphics[height=0.4in]{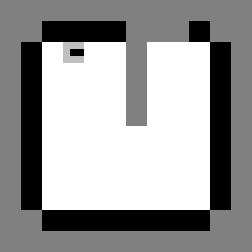}} \\
        \frame{\includegraphics[height=0.4in]{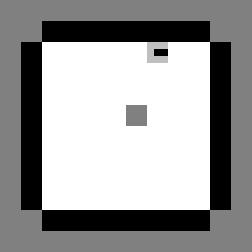}} \\
        \frame{\includegraphics[height=0.4in]{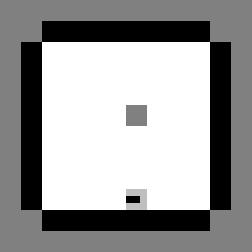}} \\
        \frame{\includegraphics[height=0.4in]{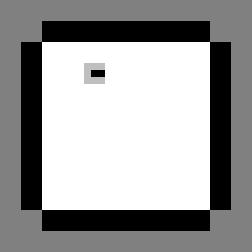}} \\
    \caption{top}
    \label{fig:memory-world}
    \end{subfigure}
    \begin{subfigure}{0.053\textwidth}
    \centering
        \frame{\includegraphics[height=0.4in]{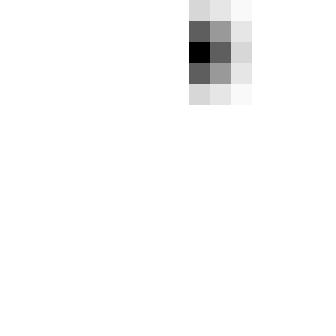}} \\
        \frame{\includegraphics[height=0.4in]{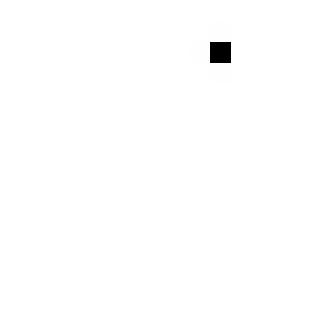}} \\
        \frame{\includegraphics[height=0.4in]{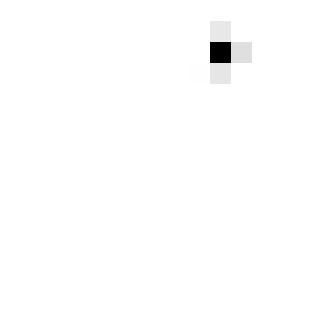}} \\
        \frame{\includegraphics[height=0.4in]{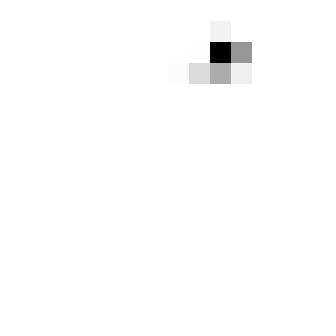}} \\
        \frame{\includegraphics[height=0.4in]{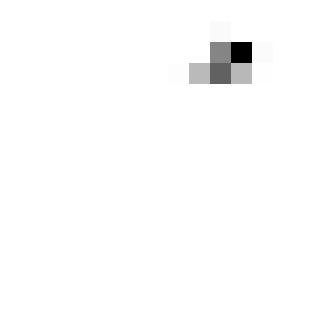}} \\
        \frame{\includegraphics[height=0.4in]{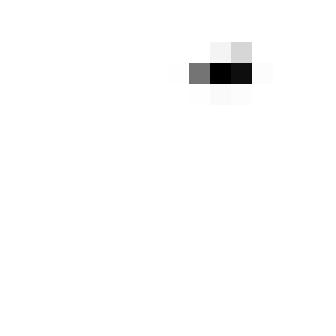}} \\
        \frame{\includegraphics[height=0.4in]{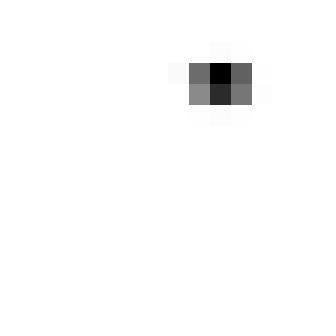}} \\
        \frame{\includegraphics[height=0.4in]{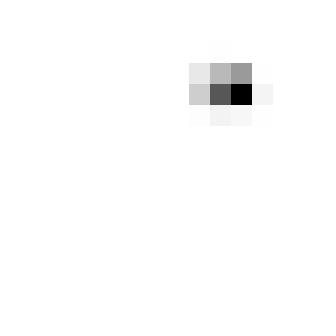}} \\
        \frame{\includegraphics[height=0.4in]{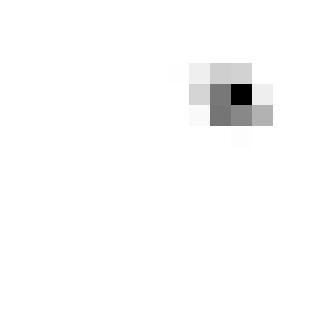}} \\
        \frame{\includegraphics[height=0.4in]{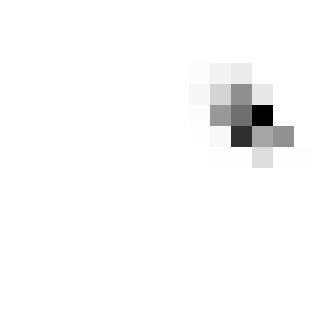}} \\
        \frame{\includegraphics[height=0.4in]{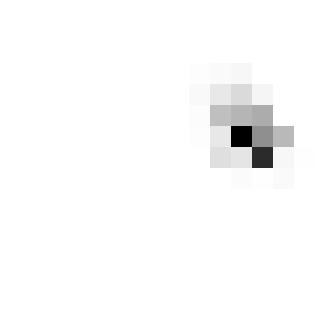}} \\
    \caption{$\bm{\omega}^{w}$}
    \label{fig:memory-write}
    \end{subfigure}
    \begin{subfigure}{0.053\textwidth}
    \centering
        \frame{\includegraphics[height=0.4in]{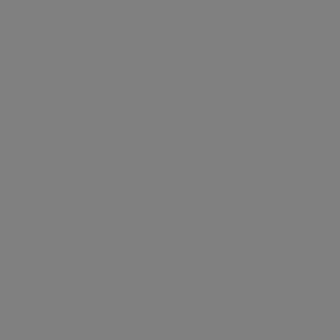}} \\
        \frame{\includegraphics[height=0.4in]{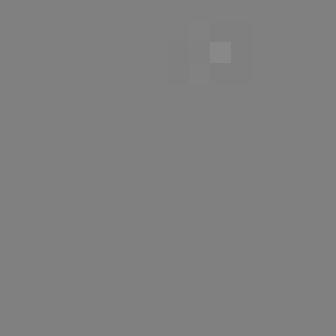}} \\
        \frame{\includegraphics[height=0.4in]{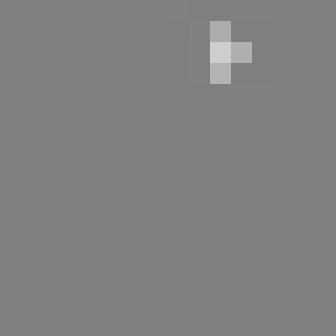}} \\
        \frame{\includegraphics[height=0.4in]{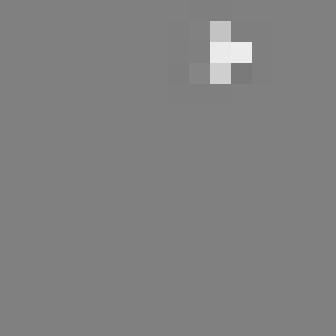}} \\
        \frame{\includegraphics[height=0.4in]{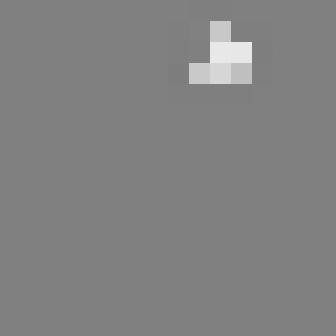}} \\
        \frame{\includegraphics[height=0.4in]{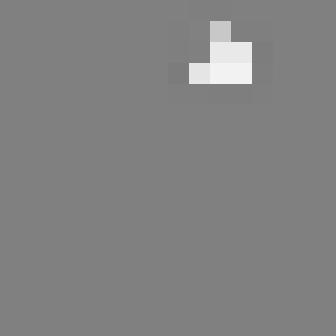}} \\
        \frame{\includegraphics[height=0.4in]{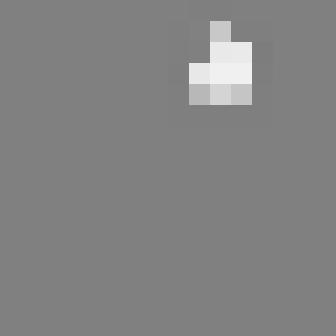}} \\
        \frame{\includegraphics[height=0.4in]{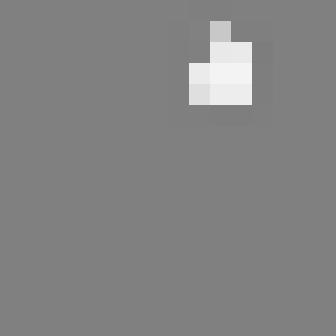}} \\
        \frame{\includegraphics[height=0.4in]{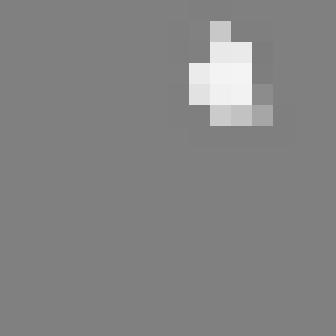}} \\
        \frame{\includegraphics[height=0.4in]{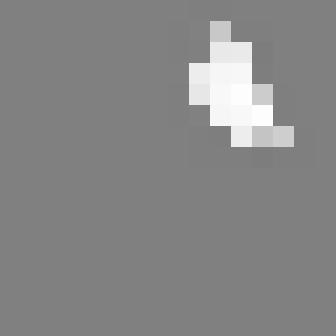}} \\
        \frame{\includegraphics[height=0.4in]{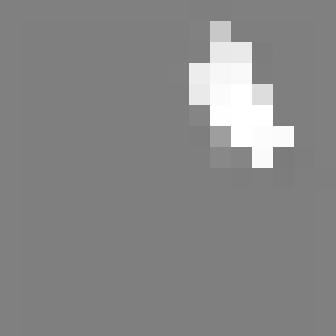}} \\
    \caption{$\mathbf{M}$}
    \label{fig:memory-memory}
    \end{subfigure}
    \begin{subfigure}{0.053\textwidth}
    \centering
        \frame{\includegraphics[height=0.4in]{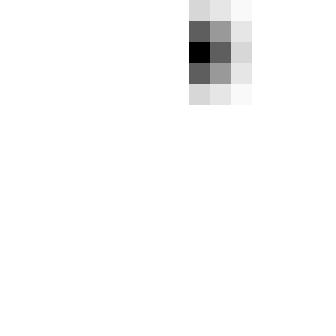}} \\
        \frame{\includegraphics[height=0.4in]{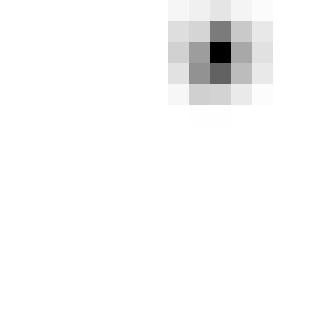}} \\
        \frame{\includegraphics[height=0.4in]{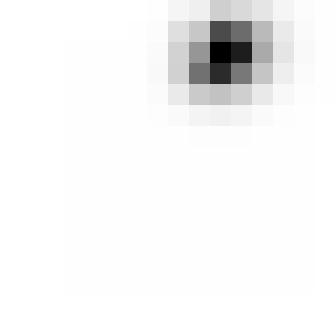}} \\
        \frame{\includegraphics[height=0.4in]{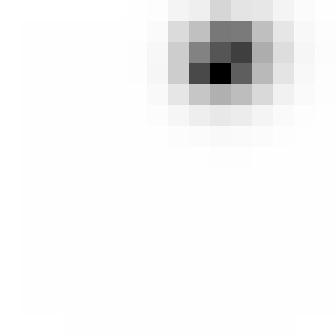}} \\
        \frame{\includegraphics[height=0.4in]{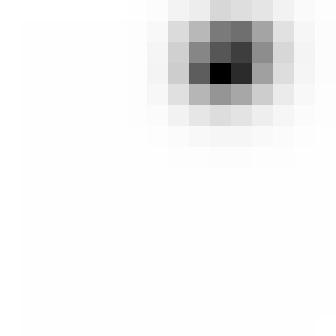}} \\
        \frame{\includegraphics[height=0.4in]{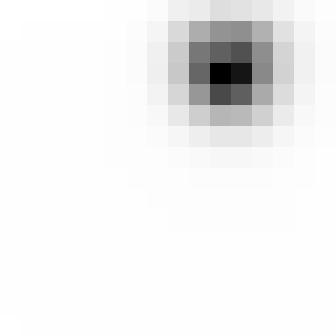}} \\
        \frame{\includegraphics[height=0.4in]{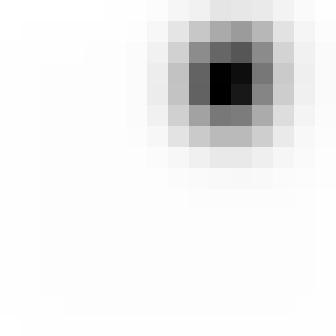}} \\
        \frame{\includegraphics[height=0.4in]{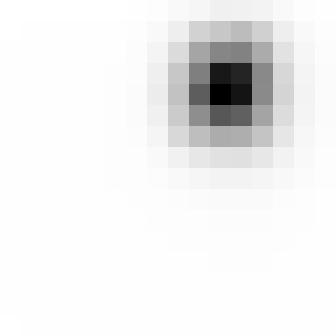}} \\
        \frame{\includegraphics[height=0.4in]{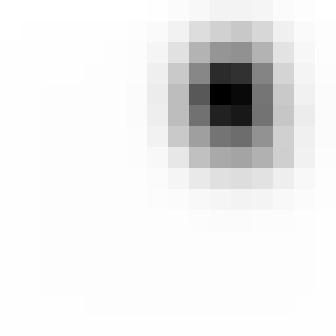}} \\
        \frame{\includegraphics[height=0.4in]{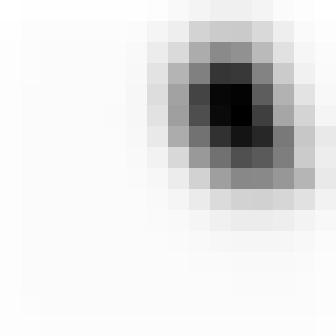}} \\
        \frame{\includegraphics[height=0.4in]{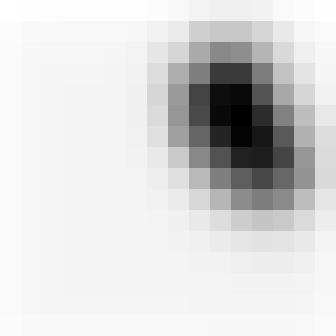}} \\
    \caption{$\bm{\omega}^{r}$}
    \label{fig:memory-read}
    \end{subfigure}
    \begin{subfigure}{0.053\textwidth}
     \centering
        \frame{\includegraphics[height=0.4in]{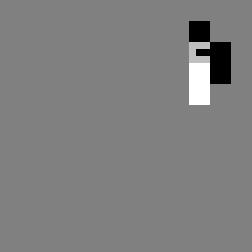}}\\
        \frame{\includegraphics[height=0.4in]{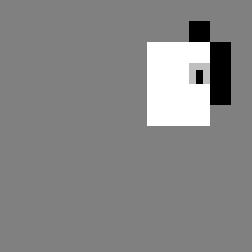}}\\
        \frame{\includegraphics[height=0.4in]{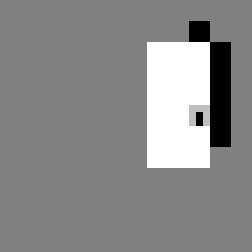}}\\
        \frame{\includegraphics[height=0.4in]{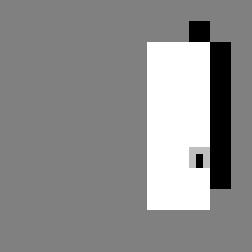}}\\
        \frame{\includegraphics[height=0.4in]{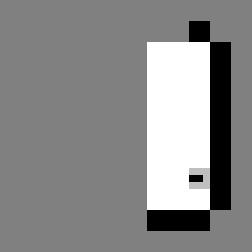}}\\
        \frame{\includegraphics[height=0.4in]{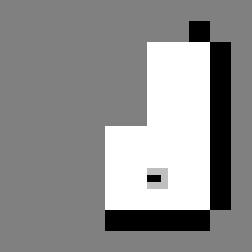}}\\
        \frame{\includegraphics[height=0.4in]{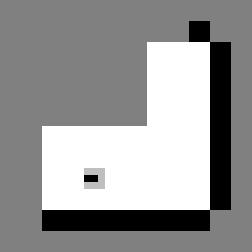}}\\
        \frame{\includegraphics[height=0.4in]{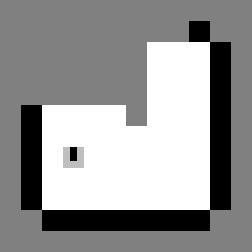}}\\
        \frame{\includegraphics[height=0.4in]{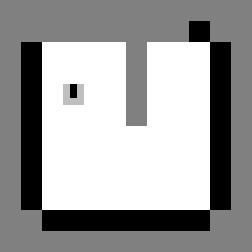}}\\
        \frame{\includegraphics[height=0.4in]{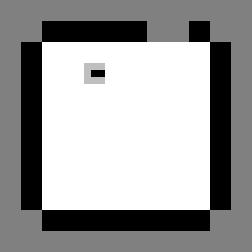}}\\
        \frame{\includegraphics[height=0.4in]{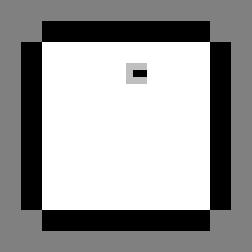}}\\
    \caption{top}
    \label{fig:memory-world-1}
    \end{subfigure}
    \begin{subfigure}{0.053\textwidth}
     \centering
        \frame{\includegraphics[height=0.4in]{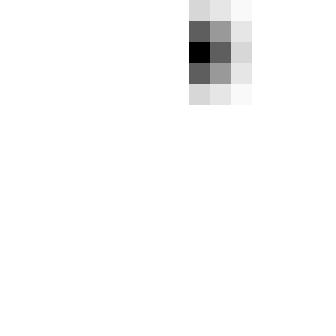}}\\
        \frame{\includegraphics[height=0.4in]{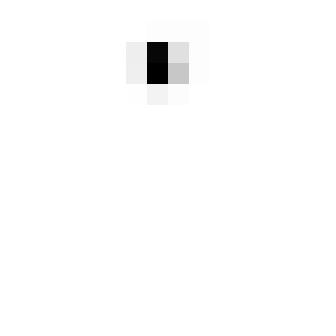}}\\
        \frame{\includegraphics[height=0.4in]{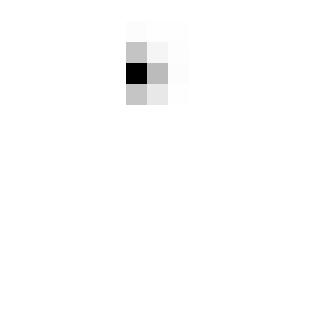}}\\
        \frame{\includegraphics[height=0.4in]{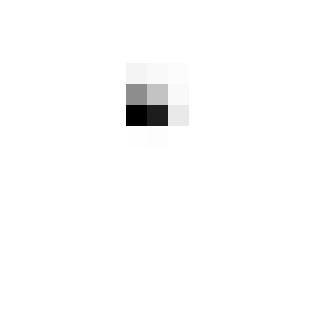}}\\
        \frame{\includegraphics[height=0.4in]{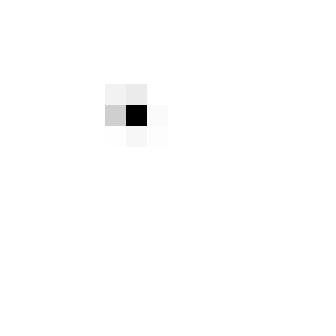}}\\
        \frame{\includegraphics[height=0.4in]{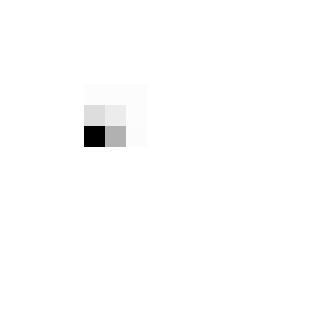}}\\
        \frame{\includegraphics[height=0.4in]{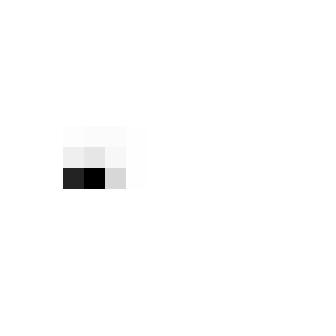}}\\
        \frame{\includegraphics[height=0.4in]{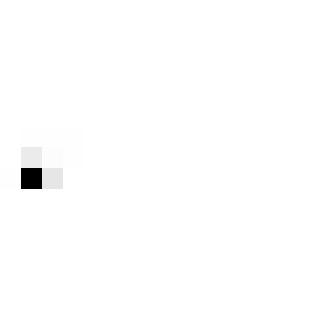}}\\
        \frame{\includegraphics[height=0.4in]{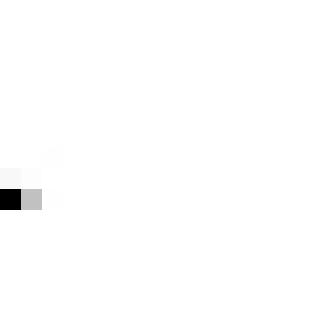}}\\
        \frame{\includegraphics[height=0.4in]{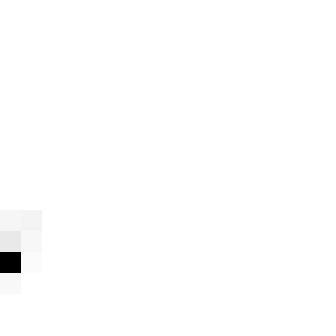}}\\
        \frame{\includegraphics[height=0.4in]{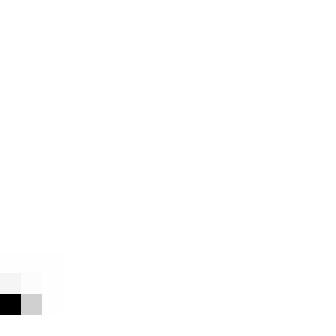}}\\
    \caption{$\bm{\omega}^{w}$}
    \label{fig:memory-write-1}
    \end{subfigure}
    \begin{subfigure}{0.053\textwidth}
     \centering
        \frame{\includegraphics[height=0.4in]{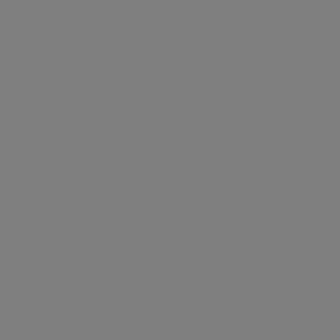}}\\
        \frame{\includegraphics[height=0.4in]{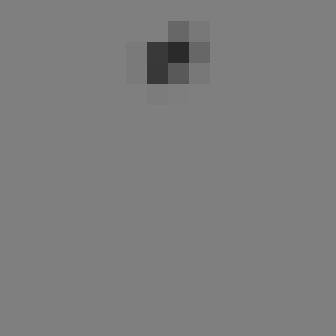}}\\
        \frame{\includegraphics[height=0.4in]{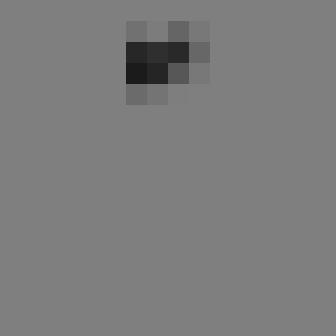}}\\
        \frame{\includegraphics[height=0.4in]{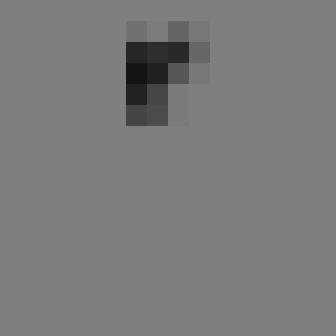}}\\
        \frame{\includegraphics[height=0.4in]{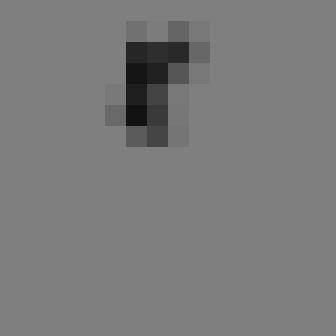}}\\
        \frame{\includegraphics[height=0.4in]{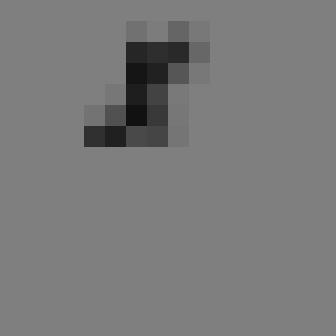}}\\
        \frame{\includegraphics[height=0.4in]{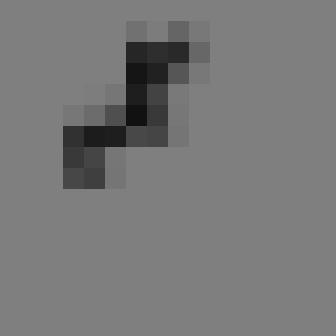}}\\
        \frame{\includegraphics[height=0.4in]{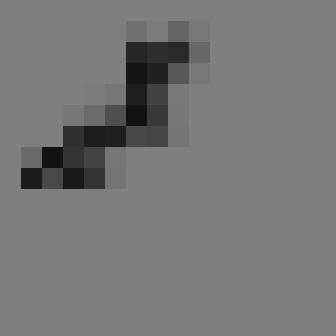}}\\
        \frame{\includegraphics[height=0.4in]{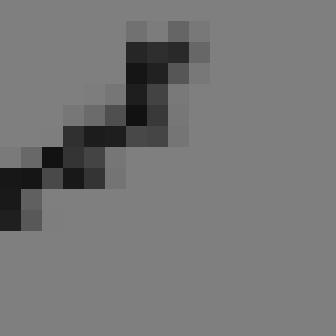}}\\
        \frame{\includegraphics[height=0.4in]{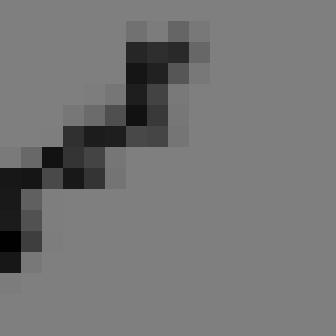}}\\
        \frame{\includegraphics[height=0.4in]{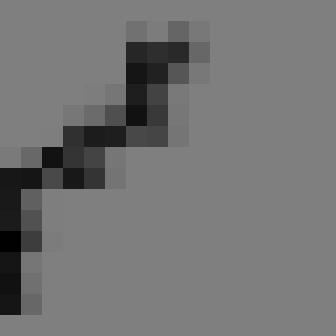}}\\
    \caption{$\mathbf{M}$}
    \label{fig:memory-memory-1}
    \end{subfigure}
    \begin{subfigure}{0.053\textwidth}
    \centering
        \frame{\includegraphics[height=0.4in]{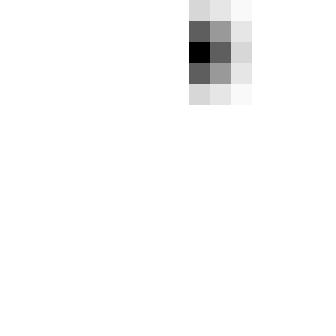}}\\
        \frame{\includegraphics[height=0.4in]{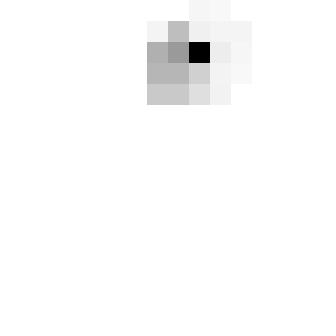}}\\
        \frame{\includegraphics[height=0.4in]{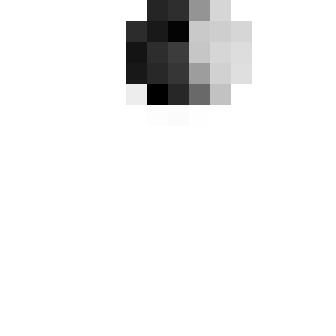}}\\
        \frame{\includegraphics[height=0.4in]{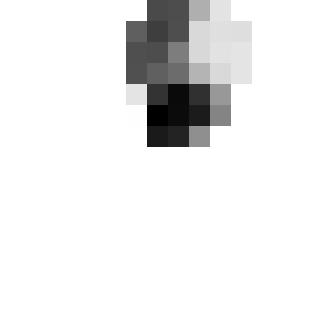}}\\
        \frame{\includegraphics[height=0.4in]{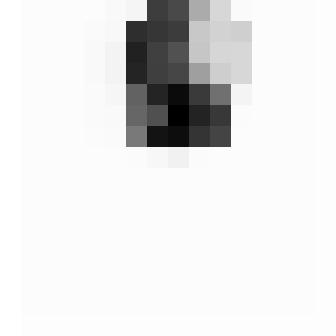}}\\
        \frame{\includegraphics[height=0.4in]{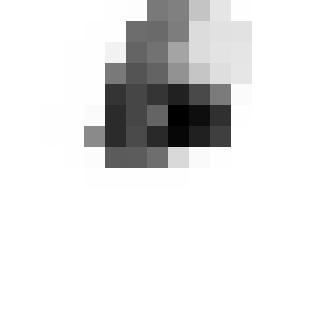}}\\
        \frame{\includegraphics[height=0.4in]{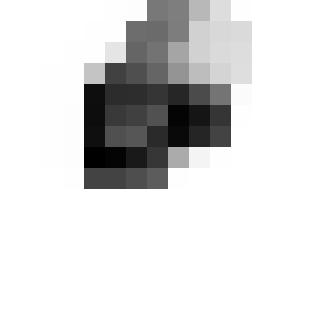}}\\
        \frame{\includegraphics[height=0.4in]{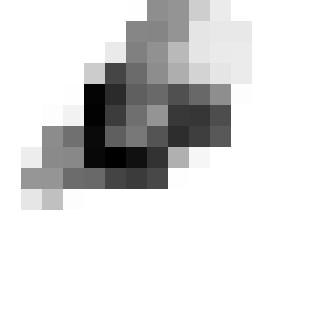}}\\
        \frame{\includegraphics[height=0.4in]{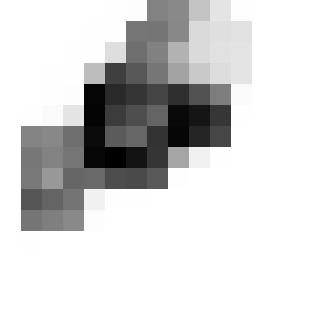}}\\
        \frame{\includegraphics[height=0.4in]{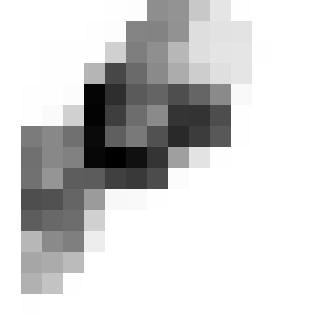}}\\
        \frame{\includegraphics[height=0.4in]{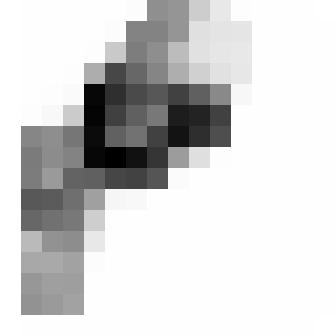}}\\
    \caption{$\bm{\omega}^{r}$}
    \label{fig:memory-read-1}
    \end{subfigure}
    \caption{
    Visualization of the top-down view,
    write head weights $\bm{\omega}^{w}$,
    external memory $\mathbf{M}$,
    and read head weights $\bm{\omega}^{r}$ during one exploration of a trained
    \textit{A3C-Ext} agent
    (Fig.\ref{fig:memory-world},\ref{fig:memory-write},\ref{fig:memory-memory},\ref{fig:memory-read}) and a trained
    \textit{Neural-SLAM} agent
    (Fig.\ref{fig:memory-world-1},\ref{fig:memory-write-1},\ref{fig:memory-memory-1},\ref{fig:memory-read-1})
    on a simple environment.
    For generating the memory visualizations Fig.\ref{fig:memory-memory},\ref{fig:memory-memory-1},
    first a sigmoid is applied on each of the
    $16\times16\times16$
    memory values,
    then the mean value taken over all the $16$ channels is visualized for for each of the
    $16\times16$
    memory slot.
    In each of the write/read head weight visualizations,
    the maxmimum value is drawn in black and the minimum in white;
    while in the memory visualizations,
    black corresponds to a value of $1$
    and white to a value of $0$.
}
    \label{fig:memory}
\end{figure}

We first test our algorithm in simulated grid world environments.
Here the agent moves in simplified dynamics,
exploring grid environments with discrete actions like turning \ang{90} or moving forward to a neighboring grid.
Considering a differential drive sweeping robot with a common size of
$30\times30\si{\centi\metre}^2$,
testing our algorithm in grid worlds of sizes up to
$16\times16$ could be seen as the sweeping robot covering rooms of size
$4.8\times4.8\si{\metre}^2$.

We use a curriculum learning strategy to train agents to explore randomly generated environments ranging from size
$8\times8$ to
$12\times12$.
We ensure all the free grids are connected in each generated environment,
several sample environments are shown in Fig.\ref{fig:maps}.
At the beginning of each episode, the agent is randomly placed in a randomly generated grid world during both training and evaluation. It has a sensing area of size $3\times5$ (we note that this simulated laser sensor cannot see through walls nor across sharp angles),
which is shown as a red bounding box in Fig.\ref{fig:demo}.
The goal of the agent is to clear up all accessible unexplored areas,
meaning those grey grids that are not blocked by the black grids which represent obstacles.
At every time step,
the agent can take an action out of
$\{
0\text{: stand still},
1\text{: }
$
$
\text{turn left by}\hspace{.1cm}\ang{90},
2\text{: turn right by}\hspace{.1cm}\ang{90},
3\text{: go forward by}\hspace{.07cm}1\hspace{.07cm}\text{grid}\}
$.
It will receive a reward of
$-0.04$
for each step it takes before completing the exploration task,
$-0.96$ for colliding with obstacles,
and $10$ for completing the coverage of the environment.
Also during the course of exploration,
the agent will receive a reward of
$\nicefrac{1}{3\times5}$
for each new grid it clears up
(we note that the calculation of this reward requires the ground truth map,
but such a map is only needed during training to provide exploration rewards,
while it will not be needed during test or execution).

As a side discussion,
we suspect that there might be the possibility of
obtaining the exploration reward
using the values of the external memory,
thus eliminating the need for the ground truth map during training.
For example,
for the experiments in this paper,
the initial values of the memory $\mathbf{M}_0$ are set to $0$ at the beginning of each episode.
Interpreting this as the case of maxmimum uncertainty,
and values that deviate far from $0$ as containing more information,
one might be able to obtain certain forms of information gain measures out of the external memory values,
which might serve as a form of intrinsic motivation
\cite{oudeyer2008can,schmidhuber2010formal}.

At each time step,
the sensing information the agent receives of size $3\times5$ will be fed into the network,
along with the action it selected from the last time step.
We use A3C \cite{mnih2016asynchronous} as the backbone deep reinforment learning algorithm
and deploy $16$ training processes purely on CPU,
optimized with the ADAM optimizer \cite{Kingma2015} with shared statistics across all training processes,
with a learning rate of $1e-4$.
We also use a weight decay of $1e-4$ since we find this to be essential to stabilize training when combining external memory architectures with A3C.
The rollout step $K$ is set to $20$ and
the maximum number of steps for each episode is $750$.

We experimented with the following baseline agents as comparisons to our \textit{Neural-SLAM} agent:
1) \textit{A3C}:
an A3C agent with one LSTM cell (with $128$ hidden units) without external memory architectures.
The action from the last time step is concatenated with the current sensor readings as the input to the network;
2) \textit{A3C-Nav1}:
an A3C agent with two stacked LSTM cells.
The last action is fed into the $1\ts{st}$ LSTM cell;
3) \textit{A3C-Nav2}:
same as \textit{A3C-Nav1} except that the last action is fed into the $2\ts{nd}$ LSTM cell
(we note that this agent is of very similar structure to the one proposed by \cite{mirowski2016learning},
while here we only feed the selected action but not the true velocity and the reward to the $2\ts{nd}$ LSTM since that information is usually not available during execution);
4) \textit{A3C-Ext}:
an A3C agent with one LSTM cell,
which also interacts with a $2D$ addressed external memory
($16\times16$ memory slots with $16$ channels each
we note that the largest map size the agents are trained on is $12\times12$,
here we initialize the address of the external memory to be $16\times16$ for the generalization tests on larger map sizes,
which will be discussed in Sec.\ref{sec:generalization}),
and access it using the same approach as we described in Sec.\ref{sec:embeddedslam}.
But unlike our \textit{Neural-SLAM} agent where the previous action is applied onto the memory through an explicit motion model,
no motion prediction step (Sec.\ref{sec:motion}) is executed for the \textit{A3C-Ext} agent,
and the previous action is simply
concatenated with the output of the LSTM
and fed to the external memory structure,
which means that this agent is not biased
or imposed to use the action
in a location/motion-related way
as the agent of \textit{Neural-SLAM}.

\subsection{Grid World Experiments}

\begin{table}[t]
   \centering
   \caption{Testing statistics for the generalization experiment, showing the performance of \textit{Random} (random actions), \textit{A3C-Nav2} (very similar to that proposed by \protect\cite{mirowski2016learning}),
   and \textit{Neural-SLAM} (ours),
   each evaluated on the same set of 50 randomly generated worlds of size 16x16.
   The maximum number of steps per episode was $750$ steps for both the \textit{A3C-Nav2} agent and our \textit{Neural-SLAM} agent.}
   \label{tab:generalizationtest}
    \resizebox{\columnwidth}{!}{%
    \begin{tabular}{c c c c}
                & Steps & Reward & Success Ratio \\
    \hline
    Random      & 5531.600 $\pm$ 4299.554 & -596.644 $\pm$ 505.436 & -     \\
    A3C         &  333.780 $\pm$  300.098 &    4.373 $\pm$  16.880 & 33/50 \\
    A3C-Nav1    &  290.500 $\pm$  275.228 &    6.938 $\pm$  15.639 & 37/50 \\
    A3C-Nav2    &  283.480 $\pm$  279.098 &    7.196 $\pm$  15.566 & 37/50 \\
    A3C-Ext     &  569.640 $\pm$  272.931 &   -8.127 $\pm$  15.408 & 18/50 \\
    Neural-SLAM &  174.920 $\pm$  174.976 &   13.732 $\pm$   9.839 & 46/50 \\
    \end{tabular}
}
\label{tab:stats}
\end{table}

We conducted experiments in simulated grid world environment,
training the baseline agents and the \textit{Neural-SLAM} agent
continuously over a curriculum of $3$ courses,
containing randomly generated grid worlds of size
$8\times8$,
$12\times12$ and
$16\times16$ respectively
(sample grid worlds are shown in Fig.\ref{fig:maps}).
The average reward obtained by all agents are shown in Fig.\ref{fig:curriculum}.
We observe that the \textit{Neural-SLAM} agent
shows a relatively consistent and stable performance
across all $3$ courses.
Specifically,
the \textit{Neural-SLAM} agent can still
successfully and reliably explore
in the $3\ts{rd}$ course
where the environments contain more
complex structures for which effective long-term memory is essential.

We visualize the memory addressing
of a trained
\textit{A3C-Ext} agent and a trained
\textit{Neural-SLAM} agent
in Fig.\ref{fig:memory}.
Both agents are placed in the same grid world environment
with the same initial pose.
The \textit{A3C-Ext} agent complets the exploration task in
$56$ steps,
and \textit{Neural-SLAM} in $24$ steps,
the frames shown in Fig.\ref{fig:memory}
are downsampled representative ones in each trajectory.
We observe that for both agents,
the write head addressing weight
$\bm{\omega}^{w}$ converges to a more focused attention,
while the read head addressing weight
$\bm{\omega}^{r}$ learns to spread out and diffuse.
We suspect that this learned behavior
could help
the resulting read vector
$\mathbf{r}$
to better summarize the current memory
for the agent to make informed planning decisions.

As discussed previously,
since the motion update as in SLAM plays an essential role for cognitive mapping,
we explicitly apply a motion prediction step in the \textit{Neural-SLAM} agent,
to bias it to incorportate the motion information into its model in a geometric transformation way.
We note that this does not mean that the resulting memory of the agent
should have a one-to-one correspondence to tranditional maps like the occupancy grid map,
as several other operations of the soft attention mechanism are executed after the motion incorporation
(Sec.\ref{sec:methods}).
In Fig.\ref{fig:memory},
we observe that the \textit{Neural-SLAM} agent
learns a more flexible and spread-out way of
mapping onto its external memory,
making more effective usage of the 2D structure of the memory slots.
As a comparison,
the \textit{A3C-Ext} agent takes the motion as input,
but no motion model is forced on it,
in the hope that the agent could learn
its own rule of performing motion update.
However,
together with the results shown in Fig.\ref{fig:curriculum},
we observe that \textit{Neural-SLAM} is more effective.
We suspect the reason for the inferior performance of \textit{A3C-Ext}
could be that, 
for solving the smaller worlds it encountered during the $1\ts{st}$ course
which contain very few complex structures, 
memory might not be not essential;
while for the more complex worlds
it encountered during the $2\ts{nd}$ and $3\ts{rd}$ courses,
which contain many dead corners and long corridors,
memory on a longer time scale is essential to enable the agent to deploy complicated exploration strategies.
This could explain its
\textit{green}-\textit{dashed} curve in Fig.\ref{fig:curriculum}:
it learns to solve the $1\ts{st}$ course relatively quickly,
but probably does not learn a very effective strategy
to use the external memory;
so in the $2\ts{nd}$ and $3\ts{rd}$ courses,
it does not adapt as quickly.
This comparison shows that embedding internal structures
as a form of inductive bias might bootstrap the agent
to identify useful behaviors
to achieve its learning goal.

We note that the memory and the weights for the write/read heads
are all initialized to the size of $16\times16$ during training,
since generalization performance tests will be conducted
on worlds of those sizes,
which will be discussed
in the following section.

\subsection{Generalization Tests}
\label{sec:generalization}

In the experiments discussed above,
the agents are trained across $3$ courses
on world sizes ranging from $8\times8$ to $12\times12$.
We conducted additional experiments
on the same set of $50$ randomly pre-generated worlds
of size $16\times16$,
each with a corresponding randomly generated starting pose,
to test the generalization capabilities of all studied agents.
In order to measure the level of difficulties of this task,
we also deploy a \textit{Random} agent
that would always select uniformly random actions.
The testing statistics for all agents
in the generalization tests
are summarized in Table \ref{tab:stats}.

\begin{figure}[b]
    \begin{subfigure}{0.24\textwidth}
    		\centering
        \includegraphics[width=.96\textwidth]{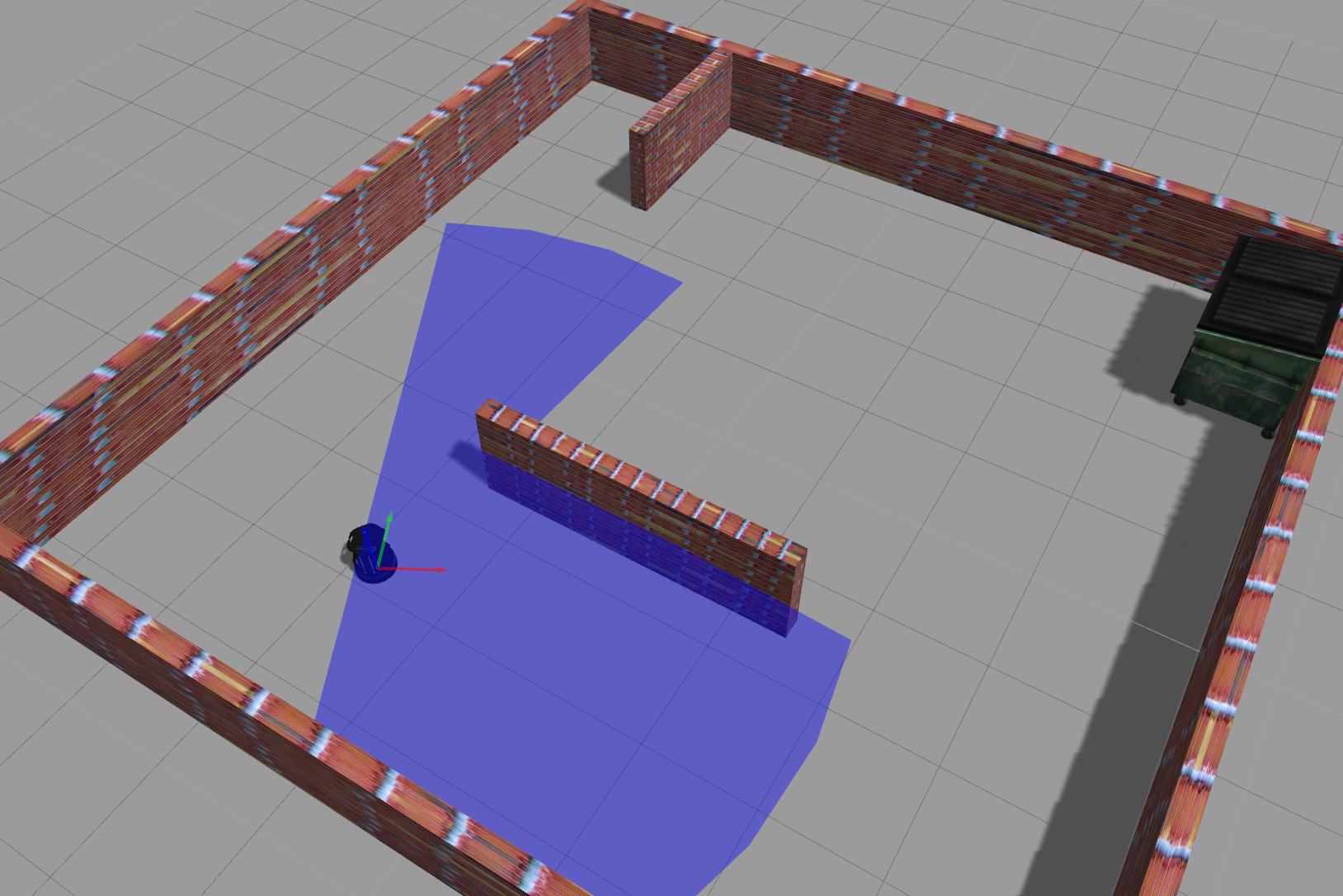}
      \vspace{.23cm}
    \end{subfigure}
    \begin{subfigure}{0.24\textwidth}
    		\centering
        \includegraphics[width=.96\textwidth]{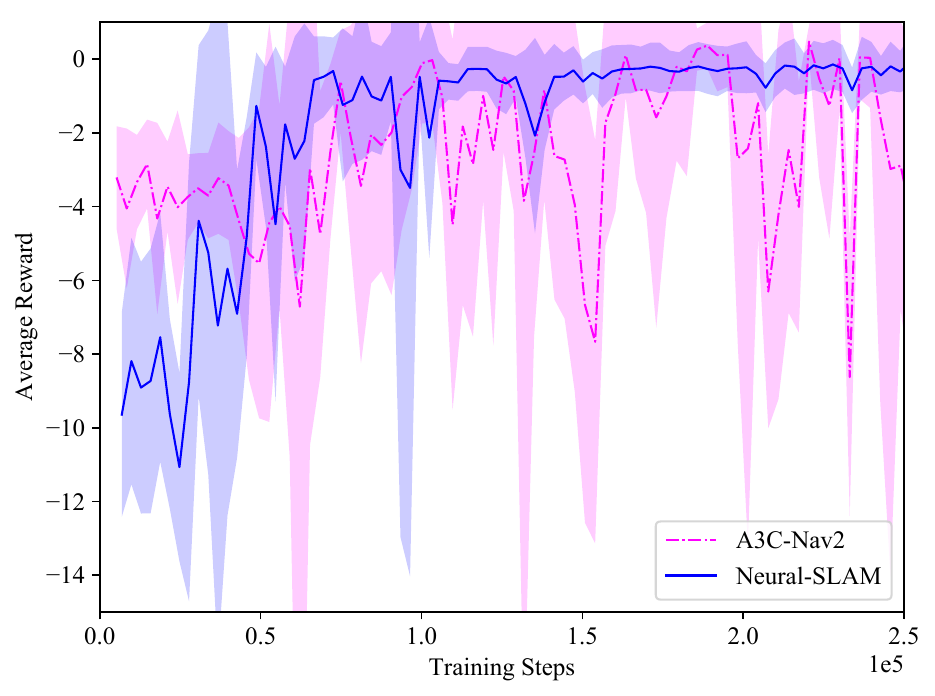}
    \end{subfigure}
    \caption{Gazebo experiments (rollout step $K$: 50; maximum steps per episode: 2500).
}
    \label{fig:gazebo}
\end{figure}

We can observe from Table \ref{tab:stats}
that the generalization task is relatively challenging,
as the \textit{Random} agent
takes an average of $5531.600$ steps to finish an episode.
We could see that the performance of \textit{A3C-Ext}
is not as satisfactory,
which could be due to reasons discussed in the last section.
We can also observe that the three agents
without access to external memory architectures
are outformed by \textit{Neural-SLAM}.
We suspect that
the lack of an external memory
to store received information about the world
could make it difficult to have access
to a global grasp of the environment,
thus planning to navigate to unexplored areas
that are far outside of the current vicinity of the agent
could be difficult for these agents.
While the \textit{Neural-SLAM} agent
is able to construct
an internal representation of the world,
which enables it to identify
and plan to go to unexplored areas
that might be relatively far away.
These different behaviors might be observed
in the supplementary video.

\subsection{Gazebo Experiments}
\label{sec:gazebo}

We also experimented with
a simple $3D$ world built in Gazebo.
We used a slightly different reward structure:
$-0.005$ as a step cost,
$-0.05$ for collision,
$1$ for the completion of an exploration,
and the exploration reward is scaled down by $0.1$
compared to the grid world experiments.
We deploy $24$ learners using docker
for training the \textit{Neural-SLAM} agent
and the $2\ts{nd}$ best performing agent
of the generalization test:
\textit{A3C-Nav2},
as training with $24$ Gazebo environments
is very computationally expensive.
From the experimental results shown in
Fig.\ref{fig:gazebo},
we can see that the \textit{Neural-SLAM} agent
is able to solve the task effectively.
We also deploy the trained agent
in new Gazebo environments
to test its generalization performance,
and we observe that the agent
is still able to accomplish exploration efficiently.
We show these experiments
in the supplementary video.



\section{Conclusions and Future Work}
\label{sec:conclusions}

We propose an approach
to provide deep reinforcement learning agents
with long-term memory capabilities,
by utilizing external memory access mechanisms.
We embed SLAM-like procedures
into the soft-attention based addressing
to bias its write/read operations
towards SLAM-like behaviors.
Our method enables the agent
to learn an internal representation
of the environment,
so as to make informative planning decisions
to effectively explore new environments.
Several interesting extensions
could emerge from our work,
including extracting internal reward signales
form the external memory
as discussed in
Sec.\ref{sec:setup}
evaluating our approach
in more challenging environments,
conducting real-world experiments,
and to experiment with higher dimensional inputs.


\bibliographystyle{plainnat}
\bibliography{paper_ns}

\end{document}